\documentclass[a4paper]{article}
\usepackage{iwslt18,amssymb,amsmath,epsfig}
\def\reg{{\rm\ooalign{\hfil
     \raise.07ex\hbox{\scriptsize R}\hfil\crcr\mathhexbox20D}}}

\usepackage{url}
\usepackage{xcolor}

\newcommand{\citet}[1]{\cite{#1}}
\newcommand{\citep}[1]{\cite{#1}}

\usepackage{graphicx}
\graphicspath{{figs/}}
\usepackage{tabularx}
\usepackage{booktabs}
\usepackage{makecell}
\usepackage{subcaption}
\usepackage{array}
\newcolumntype{L}[1]{>{\raggedright\let\newline\\\arraybackslash\hspace{0pt}}m{#1}}
\newcolumntype{C}[1]{>{\centering\let\newline\\\arraybackslash\hspace{0pt}}m{#1}}
\newcolumntype{R}[1]{>{\raggedleft\let\newline\\\arraybackslash\hspace{0pt}}m{#1}}

\usepackage{amsmath,amsmath,amsfonts,amssymb,amsthm}
\usepackage{bm}  

\newcommand{\mboshi}{Mboshi}
\newcommand{\osymb}{\ensuremath{\omega}} 
\newcommand{\wsymb}{\ensuremath{w}} 

\newcommand{\probcond}[2]{\ensuremath{P(#1\given{}#2)}}
\newcommand{\given}[0]{\ensuremath{\,\vert{}\,}}
\newcommand{\vct}[1]{\ensuremath{\boldsymbol{\mathrm{#1}}}}
\DeclareMathOperator*{\argmax}{argmax}

\def\bias{{\textsc{bias}}}
\def\aux{{\textsc{aux}}}
\def\auxratio{{\textsc{aux+ratio}}}

\def\base{{\textsc{base}}}

\newcommand{\pisa}{\texttt{Pisa}}

\title{Controlling Utterance Length in NMT-based Word Segmentation
with Attention}

 \makeatletter
 \def\name#1{\gdef\@name{#1\\}}
 \makeatother
 \name{{\em Pierre Godard$^{1,2}$, Laurent Besacier$^{2}$, Francois Yvon$^{1}$}}

 \address{(1) LIMSI, CNRS, Université Paris-Saclay, F 91 405 Orsay  \\
  (2) LIG, CNRS et Université Grenoble Alpes, F 38 000 Grenoble \\
{\small \tt pierre.godard@limsi.fr, laurent.besacier@uga.fr, francois.yvon@limsi.fr}
}
\begin{document}
\maketitle
\begin{abstract}

  One of the basic tasks of computational language documentation (CLD) is to identify
  word boundaries in an unsegmented phonemic stream. While several unsupervised
  monolingual word segmentation algorithms exist in the literature,
  they are challenged in real-world CLD settings by the small amount of available
  data. A possible remedy is to take advantage of glosses or translation in a foreign,
  well-resourced, language, which often exist for such data. In this paper, we explore and compare
  ways to exploit neural machine translation models to perform unsupervised boundary detection with bilingual information,
  notably introducing a new loss function for jointly learning alignment and segmentation. We experiment
  with an actual under-resourced language, \mboshi{}, and show that  these techniques can effectively 
  control the output segmentation length.
\end{abstract}

\section{Introduction}%
\label{sec:att:introduction}
All over the world, languages are disappearing at an unprecedented rate, fostering the need for
specific tools aimed to aid field linguists to collect, transcribe, analyze, and annotate endangered
language data (e.g.\ \cite{Anastasopoulos17spoken,Adams17phonemic}). A remarkable effort in this direction has improved the data collection procedures and tools \cite{Bird14aikuma,blachon2016parallel}, enabling to collect corpora for an increasing number of endangered languages (e.g. \cite{addaBulbSLTU2016}).

One of the basic tasks of computational language documentation (CLD) is to identify
word or morpheme boundaries in an unsegmented phonemic or orthographic stream. 
Several unsupervised monolingual word segmentation algorithms exist in the literature,
based, for instance, on information-theoretic \cite{Creutz03unsupervised,Creutz07unsupervised}
or nonparametric Bayesian techniques \cite{Goldwater06nonparametric,Johnson08unsupervised}.
These techniques are, however, challenged in real-world settings by the small amount
of available data.

A possible remedy is to take advantage of glosses or translations in a foreign,
well-resourced language (WL), which often exist for such data, hoping that the bilingual
context will provide additional cues to guide the segmentation algorithm. Such techniques
have already been explored, for instance, in \cite{Stahlberg12word,Neubig12machine} in the
context of improving statistical alignment and translation models;
and in \cite{Duong16attentional,ZanonBoito17unwritten,Godard18unsupervised} 
using Attentional
Neural Machine Translation (NMT) models. In these latter studies, word segmentation
is obtained by post-processing attention matrices, taking attention information as a
noisy proxy to word alignment \cite{Cohn16incorporating}.\footnote{This assumption is further discussed in \cite{Jain19attention} and \cite{Wiegreffe19attention}.}

In this paper, we explore ways to exploit neural machine translation models to perform unsupervised
boundary detection with bilingual information. Our main contribution is a new loss function
for jointly learning alignment and segmentation in neural translation models, allowing us to better control the length of utterances. 
Our experiments with an actual
under-resourced language (UL), Mboshi \cite{Godard18low}, show that this technique 
outperforms our bilingual 
segmentation baseline.

\section{Recurrent architectures in NMT}%
\label{sec:att:encoder_decoder_with_attention}

In this section, we briefly review the main concepts of recurrent architectures 
for machine translation introduced in \citep{Sutskever14sequence,Cho14properties,Bahdanau15neural}.\footnote{The Transformer model of \citet{Vaswani17attention} arguably yields better translation performance, but the underlying soft alignments provided by the multi-head, multi-layered attention mechanism  are harder to exploit \citet{Alkhouli18alignment,zanonboito:hal-02193867}.}  In our setting,
the source and target sentences are always observed and we are mostly interested in the attention mechanism that is used to induce word segmentation.

\subsection{RNN encoder-decoder}%
\label{sec:att:rnn_encoder_decoder}

Sequence-to-sequence models transform a variable-length \textit{source} sequence into a variable-length \textit{target} output sequence. In our context, the source sequence is a sequence of words $\wsymb_1, \ldots, \wsymb_J$ and the target sequence is an unsegmented sequence of phonemes or characters $\osymb_1, \ldots, \osymb_I$. In the RNN encoder-decoder architecture, an \textit{encoder} consisting of a RNN reads a sequence of word embeddings $e(\wsymb_1),\dots,e(\wsymb_J)$ representing the source and produces a dense representation $c$ of this sentence in a low-dimensional vector space. Vector $c$ is then fed to an RNN \textit{decoder} producing the output translation $\osymb_1,\dots,\osymb_I$ sequentially.

At each step of the input sequence, the encoder hidden states $h_j$ are computed as:
\begin{equation} \label{eq:att:encoder}
  h_j = \phi(e(\wsymb_j), h_{j-1}) \,.
\end{equation}
In most cases, $\phi$ corresponds to a long short-term memory (LSTM) \citep{Hochreiter97long} unit or a gated recurrent unit (GRU) \citep{Cho14learning}, and $h_J$ is used as the fixed-length context vector $c$ initializing the RNN decoder. 

On the target side, the decoder predicts each word $\osymb_i$, given the context vector $c$ (in the simplest case, $h_J$, the last hidden state of the encoder) and the previously predicted words, using the probability distribution over the output vocabulary $V_T$:
\begin{equation} \label{eq:att:basic_decoder_1}
  \begin{cases}
  \probcond{\osymb}{\osymb_1, \dots, \osymb_{i-1}, c} = g(\osymb_{i-1}, s_i, c) \\
  \osymb_i = \argmax_{\osymb_k} \probcond{\osymb=\osymb_k}{\osymb_1, \dots, \osymb_{i-1}, c} \,,
  \end{cases}
\end{equation}
where $s_i$ is the hidden state of the decoder RNN and $g$ is a nonlinear function (e.g. a multi-layer perceptron with a softmax layer) computed by the output layer of the decoder. The hidden state $s_i$ is then updated according to: 
\begin{equation} \label{eq:att:basic_decoder_2}
  s_i = f(s_{i-1}, e(\osymb_{i-1}), c) \,,
\end{equation}
where $f$ again corresponds to the function computed by an LSTM or GRU cell.


The encoder and the decoder are trained jointly to maximize the likelihood of the translation $\vct{\Omega}=\Omega_1, \dots, \Omega_I$ given the source sentence $\vct{w}=w_1,\dots,w_J$.
As reference target words are available during training, $\Omega_i$ (and the corresponding embedding) can be used instead of $\osymb_i$ in Equations~\eqref{eq:att:basic_decoder_1} and \eqref{eq:att:basic_decoder_2}, a technique known as \textit{teacher forcing} \cite{Williams89learning}.\footnote{As discussed below, teacher forcing will also be used at ``test'' time in our scenario, since we are not training these models for a translation task, but a word segmentation task.} 

\subsection{The attention mechanism}%
\label{sec:att:the_attention_mechanism}

Encoding a variable-length source sentence in a fixed-length vector can lead to poor translation results with long sentences \citep{Cho14properties}. To address this problem, \citet{Bahdanau15neural} introduces an attention mechanism which provides a flexible source context to better inform the decoder's decisions. This means that the fixed context vector $c$ in Equations~\eqref{eq:att:basic_decoder_1} and \eqref{eq:att:basic_decoder_2} is replaced with a position-dependent context $c_i$,
defined as: 
\begin{equation} \label{eq:att:context_vector}
  c_i = \sum_{j=1}^{J} \alpha_{ij} h_j \,,
\end{equation}
where weights $\alpha_{ij}$ are computed by an \textit{attention
  model} made of a multi-layer perceptron (MLP) followed by a softmax layer.
Denoting $a$ the function computed by the MLP, then
\begin{equation} \label{eq:att:attention_energies}
  \begin{cases}
  e_{ij} &= a(s_{i-1}, h_j) \\  
    \alpha_{ij} &= \frac{\exp(e_{ij})}{\sum_{k=1}^{J} \exp(e_{ik})} \,,
  \end{cases}
\end{equation}
where $e_{ij}$ is known as the \textit{energy} associated to $\alpha_{ij}$. Lines in the attention matrix $A = (\alpha_{ij})$ sum to 1, and weights $\alpha_{ij}$ can be interpreted as the probability that target word $\osymb_i$ is aligned to source word $\wsymb_j$.
\citep{Bahdanau15neural} qualitatively investigated such soft alignments and concluded that their model can correctly align target words to relevant source words (see also \cite{Koehn17six,Ghader17what}). Our segmentation method (Section~\ref{sec:att:word_segmentation_using_attention}) relies on the assumption that the same holds when aligning characters or phonemes on the target side to source words.

\section{Attention-based word segmentation}%
\label{sec:att:word_segmentation_using_attention}
Recall that our goal is to discover words in an unsegmented stream of target characters (or phonemes) in the under-resourced language. In this section, we first describe a baseline method inspired by the ``align to segment'' of \cite{ZanonBoito17unwritten,Godard18unsupervised}. We then propose two extensions providing the model with a signal relevant to the segmentation process, so as to move towards a joint learning of segmentation and alignment.

\subsection{Align to segment}%
\label{sec:att:align_to_segment}

An attention matrix $A = (\alpha_{ij})$ can be interpreted as a soft alignment matrix between target and source units, where each cell $\alpha_{ij}$ corresponds to the probability for target symbols $\osymb_i$ (here, a phone) to be aligned to the source word $\wsymb_j$ (cf.\ Equation~\eqref{eq:att:attention_energies}). 
In our context, where words need to be discovered on the target side, we follow \citep{ZanonBoito17unwritten,Godard18unsupervised} and perform word segmentation as follows:
\begin{enumerate}
  \item train an attentional RNN encoder-decoder model with attention using teacher forcing  (see Section~\ref{sec:att:encoder_decoder_with_attention});
\item force-decode the entire corpus and extract one attention matrix for each sentence pair.
  \item identify boundaries in the target sequences. For each target unit $\omega_i$ of the UL, we identify the source word $\wsymb_{a_i}$ to which it is most likely aligned : $\forall i, a_i = \argmax_j \alpha_{ij}$. Given these alignment links, a word segmentation is computed by introducing a word boundary in the target whenever two adjacent units are not aligned with the same source word ($a_i \neq a_{i+1}$). 
\end{enumerate}



Considering a (simulated) low-resource setting, and building on \citet{Cohn16incorporating}'s work, \citet{Duong16attentional} propose to smooth attentional alignments, either by post-processing attention matrices, or by flattening the softmax function in the attention model (see Equation~\eqref{eq:att:attention_energies}) with a temperature parameter $T$.\footnote{In this case, $\alpha_{ij} = \frac{\exp(e_{ij}/T)}{\sum_{k=1}^{J} \exp(e_{ik}/T)}$.}
This makes sense as the authors examine attentional alignments obtained while training from UL phonemes to WL words. But when translating from WL words to UL characters, this seems less useful: smoothing will encourage a character to align to many words.\footnote{A temperature below 1 would conversely sharpen the alignment distribution. We did not observe significant changes in segmentation performance varying the temperature parameter.} 
This technique is further explored by \citet{Lin18learning}, who make the temperature parameter trainable and specific to each decoding step, so that the model can learn how to control the softness or sharpness of attention distributions, depending on the current word being decoded.


\subsection{Towards joint alignment and segmentation}%
\label{sec:att:extensions}

One limitation in the approach described above lies in the absence of signal relative to segmentation during RNN training. Attempting to move towards a joint learning of alignment and segmentation, we propose here two extensions aimed at introducing constraints derived from our segmentation heuristic in the training process. 

\subsubsection{Word-length bias}%
\label{par:att:word_length_bias}


Our first extension relies on the assumption that the length of aligned source and target words should correlate. Being in a relationship of mutual translation, aligned words are expected to have comparable frequencies and meaning, hence comparable lengths.\footnote{Zipf's ``Law of Abbreviation'', a language universal, states that frequent words tend to be short. Additionally, experimental psychology also correlates word length and conceptual complexity \citep{Lewis16length}.} This means that the longer a source word is, the more target units should be aligned to it. We implement this idea in the attention mechanism as a word-length bias, changing the computation of the context vector from Equation~\eqref{eq:att:context_vector} to:
\begin{equation}
c_i = \sum_j \psi(|w_j|) \, \alpha_{ij} \, h_j
\end{equation}
where $\psi$ is a monotonically increasing function of the length $|w_j|$ of word $w_j$. This will encourage target units to attend more to longer source words.
In practice, we choose $\psi$ to be the identity function and renormalize so as to ensure that lines still sum to 1 in the attention matrices. The context vectors $c_i$ are now computed with attention weights $\tilde{\alpha}_{ij}$ as:
\begin{equation}\label{eq:att:alpha_tilde}
  \begin{cases}
    \tilde{\alpha}_{ij} &= \frac{|w_j|}{\sum_j |w_j| \, \alpha_{ij}} \, \alpha_{ij} \\
    c_i &= \sum_j \tilde{\alpha}_{ij} \, h_j \,.
  \end{cases}
\end{equation}
We finally derive the target segmentation from the attention matrix $A = (\tilde{\alpha}_{ij})$, following the method of Section~\ref{sec:att:align_to_segment}.

\subsubsection{Introducing an auxiliary loss function}%
\label{par:att:auxiliary_loss}

Another way to inject segmentation awareness inside our training procedure is to control the number of target words that will be produced during post-processing. 
The intuition here is that notwithstanding typological discrepancies, the target segmentation should yield a number of target words that is close to the length of the source.\footnote{Arguably, this constraint should better be enforced at the level of morphemes, instead of words.} 

To this end, we complement the main loss function with an additional term $\mathcal{L}_\mathrm{AUX}$ defined as:
\begin{equation}\label{eq:att:aux_loss}
\mathcal{L}_\mathrm{AUX}(\vct{\Omega}\, |\, \vct{w}) = \lvert I - J - \sum_{i=1}^{I-1} \alpha_{i,*}^\top \alpha_{i+1, *} \rvert
\end{equation}
The rationale behind this additional term is as follows: recall that a boundary is then inserted on the target side whenever two consecutive units are not aligned to the same source word. The dot product between consecutive lines in the attention matrix will be close to 1 if consecutive target units are aligned to the same source word, and closer to 0 if they are not. The summation thus quantifies the number of target units that will \textit{not} be followed by a word boundary after segmentation, and $I - \sum_{i=1}^{I-1} \alpha_{i,*}^\top \alpha_{i+1, *}$ measures the number of word boundaries that are produced on the target side. Minimizing this auxiliary term should guide the model towards learning attention matrices resulting in target segmentations that have the same number of words on the source and target sides. 

\begin{figure}[htpb]
  \centering
  \begin{subfigure}[htpb]{0.49\linewidth}
  \centering
    \includegraphics[width=0.6\linewidth]{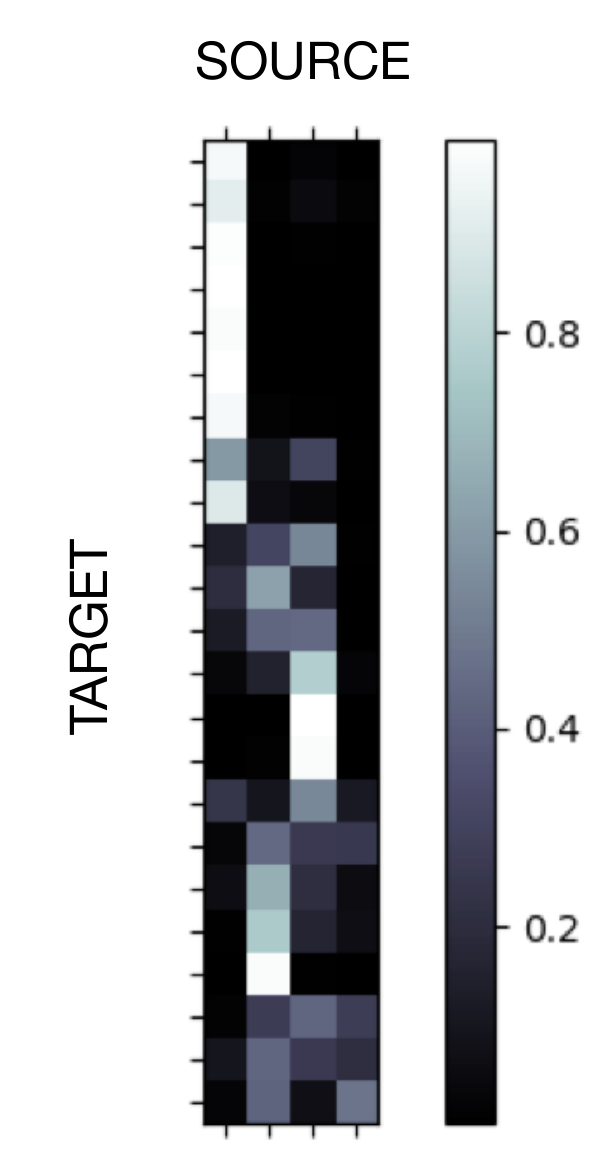}
    \caption{without auxiliary loss}
    \label{fig:att:without_aux}
  \end{subfigure}
  \begin{subfigure}[htpb]{0.49\linewidth}
  \centering
    \includegraphics[width=0.6\linewidth]{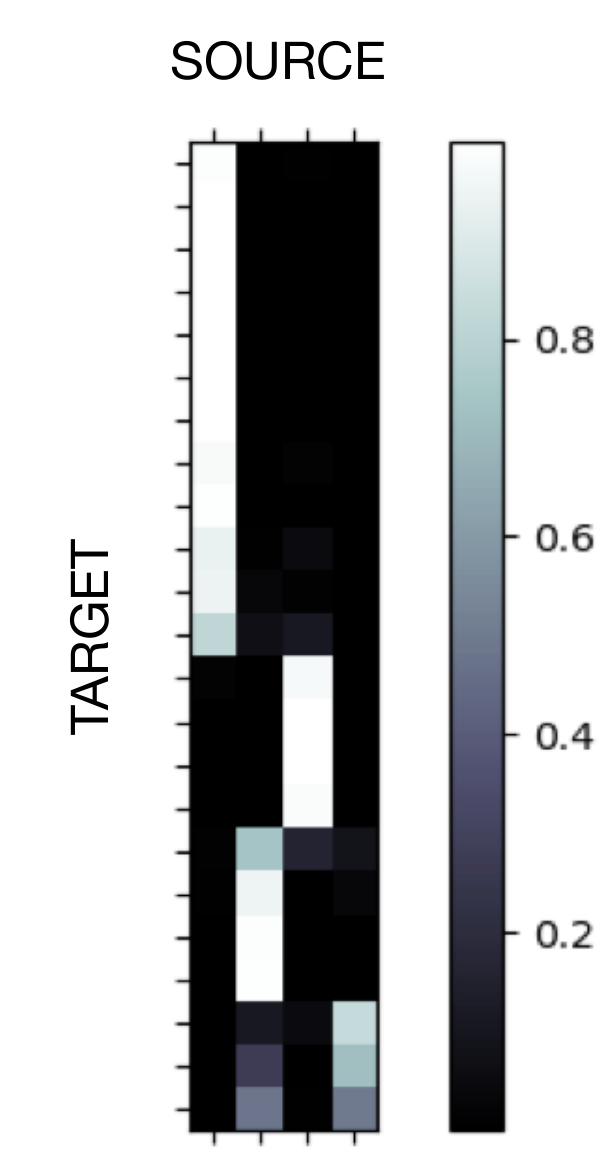}
    \caption{with auxiliary loss}
    \label{fig:att:with_aux}
  \end{subfigure}
  \caption[Effect of the auxiliary loss ($\mathcal{L}_\mathrm{NLL}$)
  on an example attention matrix.]{Effect of the
    auxiliary loss ($\mathcal{L}_\mathrm{NLL}$) on an example attention matrix for a sentence pair. Lines are indexed by target characters (or phonemes) and columns, by source words; lighter squares correspond to higher attention weights $\alpha_{ij}$.}
  \label{fig:att:auxiliary_loss}
\end{figure}
  Figure~\ref{fig:att:auxiliary_loss} illustrates the effect of our auxiliary loss on an example. Without auxiliary loss, the segmentation will yield, in this case, 8 target segments (Figure~\ref{fig:att:without_aux}), while the attention learnt with auxiliary loss will yield 5 target segments (Figure~\ref{fig:att:with_aux}); source sentence, on the other hand, has 4 tokens.\footnote{We count here the end-of-sentence token corresponding to the last column in the attention matrices.}

\section{Experiments and discussion}%
\label{sec:att:experiments_and_discussion}

In this section, we describe implementation details for our baseline segmentation system and for the extensions proposed in Section~\ref{sec:att:extensions}, before presenting data and results.

\subsection{Implementation details}%
\label{sec:att:implementation_details}

Our baseline system is our own reimplementation of Bahdanau's encoder-decoder with attention in PyTorch \citep{Paszke17automatic}.\footnote{\url{https://pytorch.org/}. We use version 0.4.1.} The last version of our code, which handles mini-batches efficiently, heavily borrows from Joost Basting's code.\footnote{\url{https://github.com/bastings/annotated_encoder_decoder}.} Source sentences include an end-of-sentence (EOS) symbol (corresponding to $w_J$ in our notation) and target sentences include both a beginning-of-sentence (BOS) and an EOS symbol. Padding of source and target sentences in mini-batches is required, as well as masking in the attention matrices and during loss computation.
Our architecture follows \citep{Bahdanau15neural} very closely with some minor changes.

\begin{description}
\item[Encoder]
  We use a single-layer bidirectional RNN \citep{Schuster97bidirectional} with GRU cells: these have been shown to perform similarly to LSTM-based RNNs \citep{Chung14empirical}, while computationally more efficient. We use 64-dimensional hidden states for the forward and backward RNNs, and for the embeddings, similarly to \citep{ZanonBoito17unwritten,Godard18unsupervised}.
  In Equation~\eqref{eq:att:encoder}, $h_j$ corresponds to the concatenation of the forward and backward states for each step $j$ of the source sequence.
\item[Attention]
  The alignment MLP model computes function $a$ from Equation~\eqref{eq:att:attention_energies} as $a(s_{i-1}, h_j)=v_a^\top \tanh(W_a s_{i-1} + U_a h_j)$ -- see Appendix A.1.2 in \citep{Bahdanau15neural} -- where $v_a$, $W_a$, and $U_a$ are weight matrices. 
  For the computation of weights $\tilde{\alpha_{ij}}$ in the word-length bias extension (Equation~\eqref{eq:att:alpha_tilde}), we arbitrarily attribute a length of 1 to the EOS symbol on the source side.     
\item[Decoder]
  The decoder is initialized using the last backward state of the encoder and a non-linear function ($\tanh$) for state $s_0$.
  We use a single-layer GRU RNN; hidden states and output embeddings are 64-dimensional. 
  In preliminary experiments, and as in \citet{Peter17generating}, we
  observed better segmentations adopting a ``generate first''
  approach during decoding, where we first generate the current target
  word, then update the current RNN state. Equations~\eqref{eq:att:basic_decoder_1} and \eqref{eq:att:basic_decoder_2} are accordingly modified into:
\begin{equation*} \label{eq:att:generate_first}
  \begin{cases}
    \probcond{\osymb_i}{\osymb_1, \dots, \osymb_{i-1}, c_i} = g(\osymb_{i-1}, s_{i-1}, c_i) \\
    s_i = f(s_{i-1}, \osymb_i, c_i) \,.
\end{cases}
\end{equation*}
During training and forced decoding, the hidden state $s_i$ is thus updated using ground-truth embeddings $e(\Omega_{i})$. $\Omega_0$ is the BOS symbol. 
Our implementation of the output layer ($g$) consists of a MLP and a
softmax.

\item[Training]
  We train for 800 epochs on the whole corpus with Adam (the learning rate is  0.001). Parameters are updated after each mini-batch of 64 sentence pairs.\footnote{Mini-batches are created anew through shuffling and length-sorting at each epoch.}
  A dropout layer \citep{Srivastava14dropout} is applied to both source and target embedding layers, with a rate of 0.5.\footnote{We also tried to add a dropout layer after the encoder and decoder RNNs (only for the ``output'' state, not the state values used inside the recursion) but this harmed our segmentation results.}
  The weights in all linear layers are initialized with Glorot's normalized method (Equation~(16) in \citep{Glorot10understanding}) and bias vectors are initialized to 0. Embeddings are initialized with the normal distribution $\mathcal{N}(0, 0.1)$.\footnote{This seemed to slightly improve segmentation results when compared to Glorot's normalized method.} Except for the bridge between the encoder and the decoder, the initialization of RNN weights is kept to PyTorch defaults.
  During training, we minimize the NLL loss $\mathcal{L}_\mathrm{NLL}$ (see Section~\ref{sec:att:rnn_encoder_decoder}), adding optionally the auxiliary loss $\mathcal{L}_\mathrm{AUX}$ (Section~\ref{par:att:auxiliary_loss}). When the auxiliary loss term is used, we schedule it to be integrated progressively so as to avoid degenerate solutions\footnote{When the NLL loss has not ``shaped'' yet the attention matrices into soft alignments, the auxiliary loss can lead to trivial optimization solutions, in which a single column in the attention matrices has a certain number of weights set to 1 (to reach the proper value in the sum term from Equation~\eqref{eq:att:aux_loss}), while all other weights in the matrices are zeroed. The model is subsequently unable to escape this solution.} with coefficient $\lambda_\mathrm{AUX}(k)$ at epoch $k$ defined by: 
\begin{equation}
  \lambda_\mathrm{AUX}(k) = \frac{\operatorname{max}(k - W)}{K}
\end{equation}
where $K$ is the total number of epochs and $W$ a \textit{wait} parameter. The complete loss at epoch $k$ is thus $\mathcal{L}_\mathrm{NLL} + \lambda_\mathrm{AUX} \cdot \mathcal{L}_\mathrm{AUX}$.  After trying values ranging from 100 to 700, we set $W$ to 200. We approximate the absolute value in Equation~\eqref{eq:att:aux_loss} by $|x| \triangleq \sqrt{x^2 + 0.001}$, in order to make the auxiliary loss function differentiable.
\end{description}

\subsection{Data and evaluation}%
\label{sec:att:data_and_evaluation}

Our experiments are performed on an actual endangered language,  {\mboshi} (Bantu C25), a language spoken in Congo-Brazzaville, using the bilingual French-Mboshi~5K corpus of \cite{Godard18low}. 
On the \mboshi{} side, we consider alphabetic representation with no tonal information. On the French side,we simply consider the default segmentation into words.\footnote{Several other units were also considered (lemmas, morphs, part-of-speech tags) but they did not lead to better performance (results not reported here). Investigation of subword units such as byte pair encodings (BPE) is left for future work.}

We denote the baseline segmentation system as \base{}, the word-length bias extension as \bias{}, and the auxiliary loss extensions as \aux{}. We also report results for a variant of \aux{} (\auxratio{}), in which the auxiliary loss is computed with a factor corresponding to the true length ratio $r_\mathrm{MB/FR}$ between \mboshi{} and French averaged over the first 100 sentences\footnote{This is a plausible supervision in the CLD scenario, and enables to relax assumption that number of words should be the same on target and source.} of the corpus. In this variant, the auxiliary loss is computed as $\lvert I - r_\mathrm{MB/FR} \cdot J - \sum_{i=1}^{I-1} \alpha_{i,*}^\top \alpha_{i+1, *} \rvert$. 

\begin{figure*}[t]
  \centering
  \includegraphics[width=1.0\textwidth]{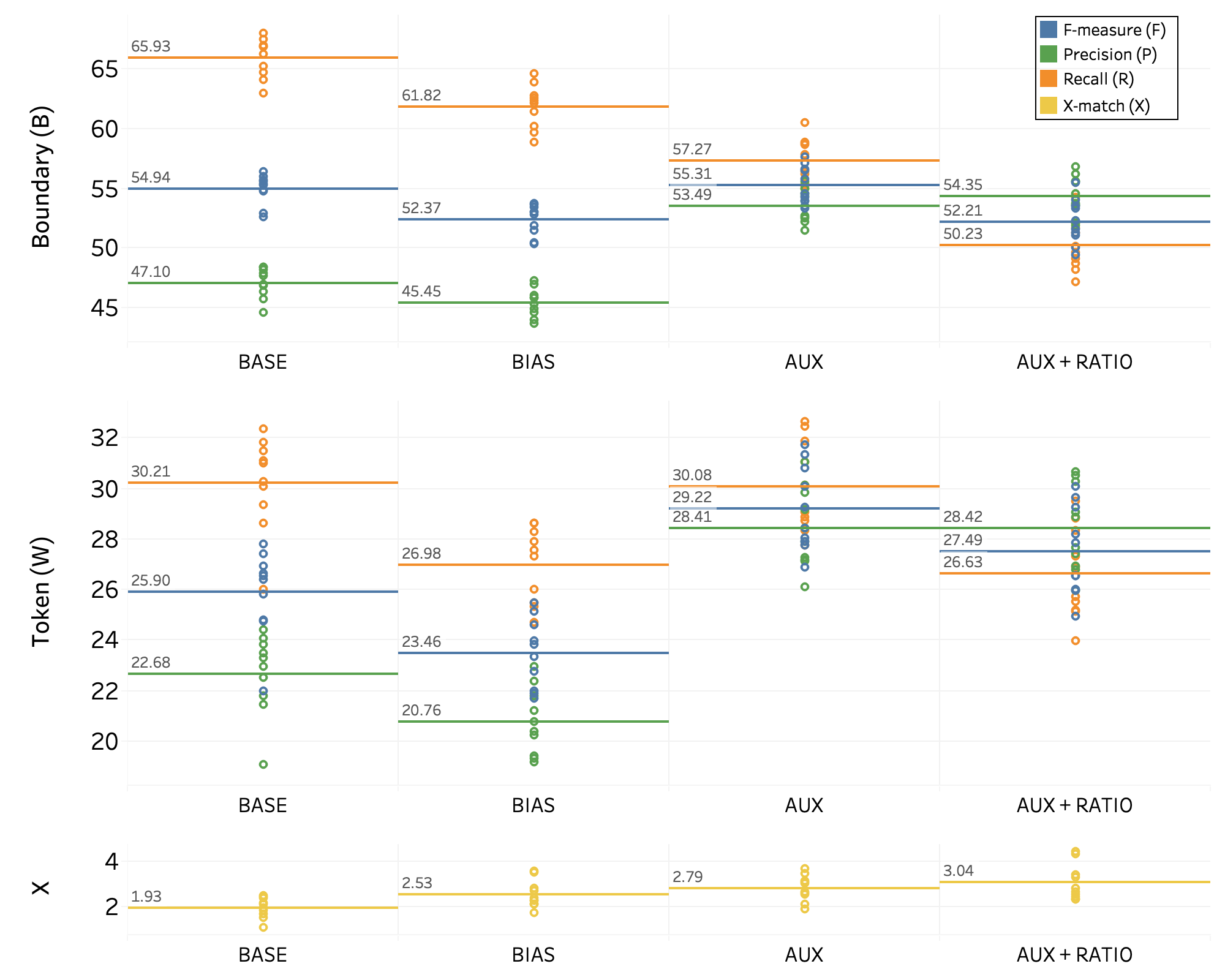}
  \caption{Boundary and token metrics (F-measure, precision, recall), and sentence exact-match (X) with methods \base{}, \bias{}, \aux{}, and \auxratio{}, on the \mboshi{} 5K corpus. Horizontal colored lines correspond to values averaged over the 10 runs.}
  \label{fig:att:base_ext}
\end{figure*}

We report segmentation performance using precision, recall, and F-measure on boundaries (BP, BR, BF), and tokens (WP, WR, WF). We also report the exact-match (X) metric which computes the proportion of correctly segmented utterances.\footnote{The exact-match metric includes single-word utterances.}
Our main results are in Figure~\ref{fig:att:base_ext},
where we report averaged scores over 10 runs.
As a comparison with another bilingual method inspired by the ``align to segment'' approach, we also include the results obtained using the statistical models of \cite{Stahlberg12word}, denoted \pisa, in Table~\ref{tab:results_pisa}.

\begin{table}[htbp]
  \centering
  \begin{tabular}{ccccccc} 
    \toprule
    BP & BR & BF & WP & WR & WF & X  \\
    \midrule
46.18 & 18.31 & 26.22 & 17.73 & 8.82 & 11.78 & 0.97 \\
    \bottomrule
  \end{tabular}
  \caption{Equivalent segmentation results with \pisa{} \cite{Stahlberg12word}.}
  \label{tab:results_pisa}
\end{table}

\subsection{Discussion}%
\label{sec:att:discussion}

A first observation is that our baseline method \base{} improves vastly over \pisa{}'s results (by a margin of about 30\% on boundary F-measure, BF).

\subsubsection{Effects of the word-length bias}%
\label{par:att:impact_of_the_word_length_bias}

The integration of a word-bias in the attention mechanism seems detrimental to segmentation performance, and results obtained with \bias{} are lower than those obtained with \base{}, except for the sentence exact-match metric (X).
To assess whether the introduction of word-length bias actually encourages target units to ``attend more'' to longer source word in \bias{}, we compute the correlation between the length of source word and the quantity of attention these words receive (for each source position, we sum attention column-wise: $\sum_i \tilde{\alpha}_{ij}$). Results for all segmentation methods are in Table~\ref{tab:att:correl}. \bias{} increases the correlation between word lengths and attention, but this correlation being already high for all methods (\base{}, or \aux{} and \auxratio{}), our attempt to increase it proves here detrimental to segmentation.

\begin{table}[htbp]
  \centering
  \small
  \begin{tabular}{C{0.09\textwidth} C{0.09\textwidth} C{0.09\textwidth} C{0.09\textwidth}} \toprule
    \base{} & \bias{} & \aux{} & \auxratio{} \\ 
    \midrule
    0.681   & \textbf{0.729}  & 0.665 & 0.662   \\
    \bottomrule
  \end{tabular}
  \caption{Correlation (avg. over 10 runs) between word length and attention (p-value for Pearson coefficient is 0 for each run) for methods \base{}, \bias{}, \aux{}, and \auxratio{}.}
  \label{tab:att:correl}
\end{table}



\subsubsection{Effects of the auxiliary loss}%
\label{par:att:impact_of_the_auxiliary_loss}

\begin{table}[htbp]
  \centering
  \small
  \begin{tabular}{lcccc} \toprule
    method & \#tokens & \#types & avg. tok. len & avg. sent. len \\ 
    \midrule
    \base{} & 40.7K & 8.2K & 3.14 & 7.94 \\
    \bias{} & 39.7K & 8.9K & 3.22 & 7.75 \\
    \aux{} & 32.3K & 9.1K & 3.95 & 6.31 \\
    \auxratio{} & 28.6K & 9.6K & 4.47 & 5.58 \\
    \midrule
    \textit{ground-truth} & \textit{30.6K} & \textit{5.3K} & \textit{4.19} & \textit{5.96} \\
    \bottomrule
  \end{tabular}
  \caption{Statistics on segmentations produced by methods \base{}, \bias{}, \aux{}, and \auxratio{}, on the \mboshi{} 5K corpus: number of tokens, types, average token length (in characters), average sentence lengths (in tokens), averaged over 10 runs.}
  \label{tab:att:base_ext_stats}
\end{table}

For boundary F-measures (BF) in Figure~\ref{fig:att:base_ext}, \aux{} performs similarly to \base{}, but with a much higher precision, and degraded recall, indicating that the new method does not oversegment as much as \base{}. More insight can be gained from various statistics on the automatically segmented data presented in Table~\ref{tab:att:base_ext_stats}.
The average token and sentence lengths for \aux{} are closer to their ground-truth values (resp. 4.19 characters and 5.96 words). The global number of tokens produced is also brought closer to its reference. On token metrics, a similar effect is observed, but the trade-off between a lower recall and an increased precision is more favorable and yields more than 3 points in F-measure. These results are encouraging for documentation purposes, where precision is arguably a more valuable metric than recall in a semi-supervised segmentation scenario.

They, however, rely on a crude heuristic that the source and target sides (here French and Mboshi) should have the same number of units, which are only valid for typologically related languages and not very accurate for our dataset. 

As \mboshi{} is more agglutinative than French (5.96 words per sentence on average in the \mboshi{} 5K, vs. 8.22 for French), we also consider the lightly supervised setting where the true length ratio is provided. This again turns out to be detrimental to performance, except for the boundary precision (BP) and the sentence exact-match (X). Note also that precision becomes stronger than recall for both boundary and token metrics, indicating under-segmentation. This is confirmed by an average token length that exceeds the ground-truth (and an average sentence length below the true value, see Table~\ref{tab:att:base_ext_stats}).

Here again, our control of the target length proves effective: compared to \base{}, the auxiliary loss has the effect to decrease the average sentence length and move it closer to its observed value (5.96), yielding an increased precision, an effect that is amplified with \auxratio{}. By tuning this ratio, it is expected that we could even get slightly better results. 



\section{Related work}
\label{sec:related}



The attention mechanism introduced by \citet{Bahdanau15neural} has been further explored by many researchers. \citet{Luong15effective}, for instance, compare a \textit{global} to a \textit{local} approach for attention, and examine several architectures to compute alignment weights $\alpha_{ij}$. 
\citet{Yang16neural} additionally propose a recurrent version of the attention mechanism, where a ``dynamic memory'' keeps track of the attention received by each source word, and demonstrate better translation results. 
A more general formulation of the attention mechanism can, lastly, be found in \citep{Kim17structured}, where structural dependencies between source units can be modeled.

With the goal of improving alignment quality, \citet{Mi16supervised} computes a distance between attentions and word alignments learnt with the reparameterization of IBM Model 2 from \citet{Dyer13simple}; this distance is then added to the cost function during training. 
To improve alignments also, \citet{Cohn16incorporating} introduce several refinements to the attention mechanism, in the form of structural biases common in word-based alignment models. 
In this work, the attention model is enriched with features able to control positional bias, fertility, or symmetry in the alignments, which leads to better translations for some language pairs, under low-resource conditions.
More work seeking to improve alignment and translation quality can be found in \citep{Tu16modeling,Liu16neural,Sankaran16temporal,Feng16implicit,Alkhouli17biasing,Kuang18attention}. 

Another important line of reseach related to work studies the relationship between segmentation and alignment quality: it is recognized that sub-lexical units such as BPE \citep{Sennrich16BPE} help solve the unknown word problem; other notable works around these lines include \cite{Kudo18subword} and \cite{Kreutzer18learning}.

CLD has also attracted a growing interest in recent years. Most recent work includes speech-to-text translation \citep{Bansal18lowresource, Bansal18pretraining}, speech transcription using bilingual supervision \citep{Anastasopoulos18leveraging}, both speech transcription and translation \citep{Anastasopoulos18tied}, or automatic phonemic transcription of tonal languages \citep{Adams18evaluating}. 

\section{Conclusion}%
\label{sec:att:conclusion}
 
In this paper, we explored neural segmentation methods extending the ``align to segment'' approach, and proposed extensions to move towards joint segmentation and alignment. This involved the introduction of a word-length bias in the attention mechanism and the design of an auxiliary loss. The latter approach yielded improvements over the baseline on all accounts, in particular for the precision metric.

Our results, however, lag behind the best monolingual performance for this dataset (see e.g.\ \cite{Godard18adaptor}). This might be due to the difficulty of computing valid alignments between phonemes and words in very limited data conditions, which remains very challenging, as also demonstrated by the results of Pisa. However, unlike monolingual methods, bilingual methods generate word alignments and their real benefit should be assessed with alignment based metrics. This is left for future work, as reference word alignments are not yet available for our data. 


Other extensions of this work will focus on ways to mitigate data sparsity with weak supervision information, either by using lists of frequent words or the presence of certain word boundaries on the target side or by using more sophisticated attention models in the spirit of \citet{Cohn16incorporating} or \citep{Kim17structured}.



%

\bibliographystyle{IEEEtran}
\bibliography{attseg}

\begin{thebibliography}{10}
\providecommand{\url}[1]{#1}
\csname url@rmstyle\endcsname
\providecommand{\newblock}{\relax}
\providecommand{\bibinfo}[2]{#2}
\providecommand\BIBentrySTDinterwordspacing{\spaceskip=0pt\relax}
\providecommand\BIBentryALTinterwordstretchfactor{4}
\providecommand\BIBentryALTinterwordspacing{\spaceskip=\fontdimen2\font plus
\BIBentryALTinterwordstretchfactor\fontdimen3\font minus
  \fontdimen4\font\relax}
\providecommand\BIBforeignlanguage[2]{{%
\expandafter\ifx\csname l@#1\endcsname\relax
\typeout{** WARNING: IEEEtran.bst: No hyphenation pattern has been}%
\typeout{** loaded for the language `#1'. Using the pattern for}%
\typeout{** the default language instead.}%
\else
\language=\csname l@#1\endcsname
\fi
#2}}

\bibitem{Anastasopoulos17spoken}
\BIBentryALTinterwordspacing
A.~Anastasopoulos, S.~Bansal, D.~Chiang, S.~Goldwater, and A.~Lopez, ``Spoken
  term discovery for language documentation using translations,'' in
  \emph{Proceedings of the Workshop on Speech-Centric Natural Language
  Processing}.\hskip 1em plus 0.5em minus 0.4em\relax Association for
  Computational Linguistics, 2017, pp. 53--58. [Online]. Available:
  \url{http://aclweb.org/anthology/W17-4607}
\BIBentrySTDinterwordspacing

\bibitem{Adams17phonemic}
\BIBentryALTinterwordspacing
O.~Adams, T.~Cohn, G.~Neubig, and A.~Michaud, ``Phonemic transcription of
  low-resource tonal languages,'' in \emph{Proceedings of the Australasian
  Language Technology Association Workshop 2017}, 2017, pp. 53--60. [Online].
  Available: \url{http://aclweb.org/anthology/U17-1006}
\BIBentrySTDinterwordspacing

\bibitem{Bird14aikuma}
S.~Bird, F.~R. Hanke, O.~Adams, and H.~Lee, ``Aikuma: A mobile app for
  collaborative language documentation,'' in \emph{Proceedings of the 2014
  Workshop on the Use of Computational Methods in the Study of Endangered
  Languages}, Baltimore, MA, 2014, pp. 1–--5.

\bibitem{blachon2016parallel}
D.~Blachon, E.~Gauthier, L.~Besacier, G.-N. Kouarata, M.~Adda-Decker, and
  A.~Rialland, ``Parallel speech collection for under-resourced language
  studies using the {LIG-Aikuma} mobile device app,'' \emph{Procedia Computer
  Science}, vol.~81, pp. 61--66, 2016.

\bibitem{addaBulbSLTU2016}
G.~Adda, S.~St\"{u}ker, M.~Adda-Decker, O.~Ambouroue, L.~Besacier, D.~Blachon,
  H.~Bonneau-Maynard, P.~Godard, F.~Hamlaoui, D.~Idiatov, G.-N. Kouarata,
  L.~Lamel, E.-M. Makasso, A.~Rialland, M.~Van~de Velde, F.~Yvon, and
  S.~Zerbian, ``Breaking the unwritten language barrier: The {Bulb} project,''
  in \emph{Proceedings of SLTU (Spoken Language Technologies for
  Under-Resourced Languages)}, Yogyakarta, Indonesia, 2016.

\bibitem{Creutz03unsupervised}
\BIBentryALTinterwordspacing
M.~Creutz, ``Unsupervised segmentation of words using prior distributions of
  morph length and frequency,'' in \emph{Proceedings of the 41st Annual Meeting
  of the Association for Computational Linguistics}.\hskip 1em plus 0.5em minus
  0.4em\relax Sapporo, Japan: Association for Computational Linguistics, July
  2003, pp. 280--287. [Online]. Available:
  \url{http://www.aclweb.org/anthology/P03-1036}
\BIBentrySTDinterwordspacing

\bibitem{Creutz07unsupervised}
\BIBentryALTinterwordspacing
M.~Creutz and K.~Lagus, ``Unsupervised models for morpheme segmentation and
  morphology learning,'' \emph{ACM Trans. Speech Lang. Process.}, vol.~4,
  no.~1, pp. 3:1--3:34, Feb. 2007. [Online]. Available:
  \url{http://doi.acm.org/10.1145/1187415.1187418}
\BIBentrySTDinterwordspacing

\bibitem{Goldwater06nonparametric}
S.~Goldwater, ``Nonparametric {{Bayesian}} models of lexical acquisition,''
  Ph.D. dissertation, Brown University, 2006.

\bibitem{Johnson08unsupervised}
M.~Johnson, ``Unsupervised {{Word Segmentation}} for {{Sesotho Using Adaptor
  Grammars}},'' in \emph{Proceedings of the {{Tenth Meeting}} of {{ACL Special
  Interest Group}} on {{Computational Morphology}} and {{Phonology}}}.\hskip
  1em plus 0.5em minus 0.4em\relax Columbus, Ohio: {Association for
  Computational Linguistics}, 2008, pp. 20--27.

\bibitem{Stahlberg12word}
F.~Stahlberg, T.~Schlippe, S.~Vogel, and T.~Schultz, ``Word segmentation
  through cross-lingual word-to-phoneme alignment,'' in \emph{Spoken {{Language
  Technology Workshop}} ({{SLT}}), 2012 {{IEEE}}}.\hskip 1em plus 0.5em minus
  0.4em\relax {IEEE}, 2012, pp. 85--90.

\bibitem{Neubig12machine}
\BIBentryALTinterwordspacing
G.~Neubig, T.~Watanabe, S.~Mori, and T.~Kawahara, ``Machine translation without
  words through substring alignment,'' in \emph{Proceedings of the 50th Annual
  Meeting of the Association for Computational Linguistics (Volume 1: Long
  Papers)}.\hskip 1em plus 0.5em minus 0.4em\relax Jeju Island, Korea:
  Association for Computational Linguistics, July 2012, pp. 165--174. [Online].
  Available: \url{http://www.aclweb.org/anthology/P12-1018}
\BIBentrySTDinterwordspacing

\bibitem{Duong16attentional}
L.~Duong, A.~Anastasopoulos, D.~Chiang, S.~Bird, and T.~Cohn, ``An attentional
  model for speech translation without transcription,'' in \emph{Proceedings of
  the 2016 Conference of the North {A}merican Chapter of the Association for
  Computational Linguistics: Human Language Technologies}.\hskip 1em plus 0.5em
  minus 0.4em\relax San Diego, California: Association for Computational
  Linguistics, June 2016, pp. 949--959.

\bibitem{ZanonBoito17unwritten}
M.~Zanon~Boito, A.~B\'erard, A.~Villavicencio, and L.~Besacier, ``Unwritten
  {{Languages Demand Attention Too}}! {{Word Discovery}} with
  {{Encoder}}-{{Decoder Models}},'' in \emph{Automatic {{Speech Recognition}}
  and {{Understanding}} ({{ASRU}}), 2017 {{IEEE Workshop}} On}.\hskip 1em plus
  0.5em minus 0.4em\relax {IEEE}, 2017.

\bibitem{Godard18unsupervised}
P.~Godard, M.~Zanon~Boito, L.~Ondel, A.~B\'erard, F.~Yvon, A.~Villavicencio,
  and L.~Besacier, ``Unsupervised {{Word Segmentation}} from {{Speech}} with
  {{Attention}},'' in \emph{Proceedings of {{Interspeech}}}, Hyderabad, India,
  2018.

\bibitem{Cohn16incorporating}
T.~Cohn, C.~D.~V. Hoang, E.~Vymolova, K.~Yao, C.~Dyer, and G.~Haffari,
  ``Incorporating structural alignment biases into an attentional neural
  translation model,'' in \emph{Proceedings of the 2016 Conference of the North
  {A}merican Chapter of the Association for Computational Linguistics: Human
  Language Technologies}.\hskip 1em plus 0.5em minus 0.4em\relax San Diego,
  California: Association for Computational Linguistics, June 2016, pp.
  876--885.

\bibitem{Jain19attention}
S.~Jain and B.~C. Wallace, ``{A}ttention is not {E}xplanation,'' in
  \emph{Proceedings of the 2019 Conference of the North {A}merican Chapter of
  the Association for Computational Linguistics: Human Language Technologies,
  Volume 1 (Long and Short Papers)}.\hskip 1em plus 0.5em minus 0.4em\relax
  Minneapolis, Minnesota: Association for Computational Linguistics, June 2019,
  pp. 3543--3556.

\bibitem{Wiegreffe19attention}
S.~Wiegreffe and Y.~Pinter, ``Attention is not not explanation,'' 2019.

\bibitem{Godard18low}
P.~Godard, G.~Adda, M.~Adda-Decker, J.~Benjumea, L.~Besacier,
  J.~Cooper-Leavitt, G.~Kouarata, L.~Lamel, H.~Maynard, M.~M{\"{u}}ller,
  A.~Rialland, S.~St{\"{u}}ker, F.~Yvon, and M.~Z. Boito,
  ``\BIBforeignlanguage{english}{{A Very Low Resource Language Speech Corpus
  for Computational Language Documentation Experiments}},'' in
  \emph{\BIBforeignlanguage{english}{Proceedings of the Language Resource and
  Evaluation Conference}}, Miyazaki, Japan, 2018.

\bibitem{Sutskever14sequence}
I.~Sutskever, O.~Vinyals, and Q.~V. Le, ``Sequence to sequence learning with
  neural networks,'' in \emph{Advances in {{Neural Information Processing
  Systems}}}, 2014, pp. 3104--3112.

\bibitem{Cho14properties}
K.~Cho, B.~{van Merrienboer}, D.~Bahdanau, and Y.~Bengio, ``On the
  {{Properties}} of {{Neural Machine Translation}}:
  {{Encoder}}\textendash{{Decoder Approaches}},'' in \emph{Proceedings of
  {{SSST}}-8, {{Eighth Workshop}} on {{Syntax}}, {{Semantics}} and
  {{Structure}} in {{Statistical Translation}}}.\hskip 1em plus 0.5em minus
  0.4em\relax Doha, Qatar: {Association for Computational Linguistics}, Oct.
  2014, pp. 103--111.

\bibitem{Bahdanau15neural}
D.~Bahdanau, K.~Cho, and Y.~Bengio, ``Neural machine translation by jointly
  learning to align and translate,'' in \emph{Proceedings of the first
  International Conference on Learning Representations}, ser. ICLR 2015, San
  Diego, CA, 2015.

\bibitem{Vaswani17attention}
A.~Vaswani, N.~Shazeer, N.~Parmar, J.~Uszkoreit, L.~Jones, A.~N. Gomez,
  L.~Kaiser, and I.~Polosukhin, ``Attention {{Is All You Need}},''
  \emph{arXiv:1706.03762 [cs]}, June 2017.

\bibitem{Alkhouli18alignment}
T.~Alkhouli, G.~Bretschner, and H.~Ney, ``On {{The Alignment Problem In
  Multi}}-{{Head Attention}}-{{Based Neural Machine Translation}},'' in
  \emph{Proceedings of the {{Third Conference}} on {{Machine Translation}}:
  {{Research Papers}}}.\hskip 1em plus 0.5em minus 0.4em\relax Belgium,
  Brussels: {Association for Computational Linguistics}, Oct. 2018, pp.
  177--185.

\bibitem{zanonboito:hal-02193867}
\BIBentryALTinterwordspacing
M.~Zanon~Boito, A.~Villavicencio, and L.~Besacier, ``{Empirical Evaluation of
  Sequence-to-Sequence Models for Word Discovery in Low-resource Settings},''
  in \emph{{Interspeech 2019}}, Graz, Austria, Sept. 2019. [Online]. Available:
  \url{https://hal.archives-ouvertes.fr/hal-02193867}
\BIBentrySTDinterwordspacing

\bibitem{Hochreiter97long}
S.~Hochreiter and J.~Schmidhuber, ``Long {{Short}}-{{Term Memory}},''
  \emph{Neural Computation}, vol.~9, no.~8, pp. 1735--1780, Nov. 1997.

\bibitem{Cho14learning}
K.~Cho, B.~{van Merrienboer}, C.~Gulcehre, D.~Bahdanau, F.~Bougares,
  H.~Schwenk, and Y.~Bengio, ``Learning {{Phrase Representations}} using {{RNN
  Encoder}}\textendash{{Decoder}} for {{Statistical Machine Translation}},'' in
  \emph{Proceedings of the 2014 {{Conference}} on {{Empirical Methods}} in
  {{Natural Language Processing}} ({{EMNLP}})}.\hskip 1em plus 0.5em minus
  0.4em\relax Doha, Qatar: {Association for Computational Linguistics}, Oct.
  2014, pp. 1724--1734.

\bibitem{Williams89learning}
R.~J. Williams and D.~Zipser, ``A {{Learning Algorithm}} for {{Continually
  Running Fully Recurrent Neural Networks}},'' \emph{Neural Computation},
  vol.~1, no.~2, pp. 270--280, June 1989.

\bibitem{Koehn17six}
P.~Koehn and R.~Knowles, ``Six {{Challenges}} for {{Neural Machine
  Translation}},'' in \emph{Proceedings of the {{First Workshop}} on {{Neural
  Machine Translation}}}.\hskip 1em plus 0.5em minus 0.4em\relax Vancouver:
  {Association for Computational Linguistics}, Aug. 2017, pp. 28--39.

\bibitem{Ghader17what}
H.~Ghader and C.~Monz, ``What does {{Attention}} in {{Neural Machine
  Translation Pay Attention}} to?'' \emph{arXiv:1710.03348 [cs]}, Oct. 2017.

\bibitem{Lin18learning}
J.~Lin, X.~Sun, X.~Ren, M.~Li, and Q.~Su, ``Learning when to concentrate or
  divert attention: Self-adaptive attention temperature for neural machine
  translation,'' in \emph{Proceedings of the 2018 Conference on Empirical
  Methods in Natural Language Processing}, ser. EMNLP 2018.\hskip 1em plus
  0.5em minus 0.4em\relax Brussels, Belgium: Association for Computational
  Linguistics, 2018, pp. 2985--2990.

\bibitem{Lewis16length}
M.~L. Lewis and M.~C. Frank, ``\BIBforeignlanguage{eng}{The length of words
  reflects their conceptual complexity},''
  \emph{\BIBforeignlanguage{eng}{Cognition}}, vol. 153, pp. 182--195, Aug.
  2016.

\bibitem{Paszke17automatic}
A.~Paszke, S.~Gross, S.~Chintala, G.~Chanan, E.~Yang, Z.~DeVito, Z.~Lin,
  A.~Desmaison, L.~Antiga, and A.~Lerer, ``Automatic differentiation in
  {{PyTorch}},'' in \emph{{{NIPS}}-{{W}}}, 2017.

\bibitem{Schuster97bidirectional}
M.~Schuster and K.~Paliwal, ``Bidirectional {{Recurrent Neural Networks}},''
  \emph{Trans. Sig. Proc.}, vol.~45, no.~11, pp. 2673--2681, Nov. 1997.

\bibitem{Chung14empirical}
J.~Chung, C.~Gulcehre, K.~Cho, and Y.~Bengio, ``\BIBforeignlanguage{English
  (US)}{{Empirical evaluation of gated recurrent neural networks on sequence
  modeling}},'' in \emph{\BIBforeignlanguage{English (US)}{{NIPS 2014 Workshop
  on Deep Learning, December 2014}}}, 2014.

\bibitem{Peter17generating}
J.-T. Peter, A.~Nix, and H.~Ney, ``\BIBforeignlanguage{en}{Generating
  {{Alignments Using Target Foresight}} in {{Attention}}-{{Based Neural Machine
  Translation}}},'' \emph{\BIBforeignlanguage{en}{The Prague Bulletin of
  Mathematical Linguistics}}, vol. 108, no.~1, pp. 27--36, June 2017.

\bibitem{Srivastava14dropout}
N.~Srivastava, G.~Hinton, A.~Krizhevsky, I.~Sutskever, and R.~Salakhutdinov,
  ``Dropout: {{A Simple Way}} to {{Prevent Neural Networks}} from
  {{Overfitting}},'' \emph{The Journal of Machine Learning Research}, vol.~15,
  no.~1, pp. 1929--1958, Jan. 2014.

\bibitem{Glorot10understanding}
X.~Glorot and Y.~Bengio, ``Understanding the difficulty of training deep
  feedforward neural networks,'' in \emph{Proceedings of the {{Thirteenth
  International Conference}} on {{Artificial Intelligence}} and
  {{Statistics}}}, ser. Proceedings of {{Machine Learning Research}}, Y.~W. Teh
  and M.~Titterington, Eds., vol.~9.\hskip 1em plus 0.5em minus 0.4em\relax
  Chia Laguna Resort, Sardinia, Italy: {PMLR}, May 2010, pp. 249--256.

\bibitem{Luong15effective}
T.~Luong, H.~Pham, and C.~D. Manning, ``Effective {{Approaches}} to
  {{Attention}}-based {{Neural Machine Translation}},'' \emph{Proceedings of
  the 2015 Conference on Empirical Methods in Natural Language Processing}, pp.
  1412--1421, 2015.

\bibitem{Yang16neural}
Z.~Yang, Z.~Hu, Y.~Deng, C.~Dyer, and A.~Smola, ``Neural {{Machine
  Translation}} with {{Recurrent Attention Modeling}},'' \emph{arXiv preprint
  arXiv:1607.05108}, 2016.

\bibitem{Kim17structured}
Y.~Kim, C.~Denton, L.~Hoang, and A.~M. Rush,
  ``\BIBforeignlanguage{en}{Structured {{Attention Networks}}},'' in
  \emph{\BIBforeignlanguage{en}{5th {{International Conference}} on {{Learning
  Representations}}}}, 2017, p.~21.

\bibitem{Mi16supervised}
H.~Mi, Z.~Wang, and A.~Ittycheriah, ``Supervised {{Attentions}} for {{Neural
  Machine Translation}},'' in \emph{Proceedings of the 2016 {{Conference}} on
  {{Empirical Methods}} in {{Natural Language Processing}}}.\hskip 1em plus
  0.5em minus 0.4em\relax Austin, Texas: {Association for Computational
  Linguistics}, Nov. 2016, pp. 2283--2288.

\bibitem{Dyer13simple}
C.~Dyer, V.~Chahuneau, and N.~A. Smith, ``A {{Simple}}, {{Fast}}, and
  {{Effective Reparameterization}} of {{IBM Model}} 2,'' in \emph{Proceedings
  of the 2013 {{Conference}} of the {{North American Chapter}} of the
  {{Association}} for {{Computational Linguistics}}: {{Human Language
  Technologies}}}.\hskip 1em plus 0.5em minus 0.4em\relax Atlanta, Georgia:
  {Association for Computational Linguistics}, June 2013, pp. 644--648.

\bibitem{Tu16modeling}
Z.~Tu, Z.~Lu, Y.~Liu, X.~Liu, and H.~Li, ``Modeling {{Coverage}} for {{Neural
  Machine Translation}},'' in \emph{Proceedings of the 54th {{Annual Meeting}}
  of the {{Association}} for {{Computational Linguistics}} ({{Volume}} 1:
  {{Long Papers}})}.\hskip 1em plus 0.5em minus 0.4em\relax Berlin, Germany:
  {Association for Computational Linguistics}, Aug. 2016, pp. 76--85.

\bibitem{Liu16neural}
L.~Liu, M.~Utiyama, A.~Finch, and E.~Sumita, ``Neural machine translation with
  supervised attention,'' in \emph{Proceedings of the 26th International
  Conference on Computational Linguistics: Technical Papers}, ser. COLING
  2016.\hskip 1em plus 0.5em minus 0.4em\relax Osaka, Japan: The COLING 2016
  Organizing Committee, 2016, pp. 3093--3102.

\bibitem{Sankaran16temporal}
B.~Sankaran, H.~Mi, Y.~{Al-Onaizan}, and A.~Ittycheriah,
  ``\BIBforeignlanguage{en}{Temporal {{Attention Model}} for {{Neural Machine
  Translation}}},'' Aug. 2016.

\bibitem{Feng16implicit}
S.~Feng, S.~Liu, M.~Li, and M.~Zhou, ``\BIBforeignlanguage{en}{Implicit
  {{Distortion}} and {{Fertility Models}} for {{Attention}}-based
  {{Encoder}}-{{Decoder NMT Model}}},'' Jan. 2016.

\bibitem{Alkhouli17biasing}
T.~Alkhouli and H.~Ney, ``Biasing {{Attention}}-{{Based Recurrent Neural
  Networks Using External Alignment Information}},'' in \emph{Proceedings of
  the {{Second Conference}} on {{Machine Translation}}}, 2017, pp. 108--117.

\bibitem{Kuang18attention}
\BIBentryALTinterwordspacing
S.~Kuang, J.~Li, A.~Branco, W.~Luo, and D.~Xiong, ``Attention focusing for
  neural machine translation by bridging source and target embeddings,'' in
  \emph{Proceedings of the 56th Annual Meeting of the Association for
  Computational Linguistics (Volume 1: Long Papers)}.\hskip 1em plus 0.5em
  minus 0.4em\relax Melbourne, Australia: Association for Computational
  Linguistics, July 2018, pp. 1767--1776. [Online]. Available:
  \url{https://www.aclweb.org/anthology/P18-1164}
\BIBentrySTDinterwordspacing

\bibitem{Sennrich16BPE}
\BIBentryALTinterwordspacing
R.~Sennrich, B.~Haddow, and A.~Birch, ``Neural machine translation of rare
  words with subword units,'' in \emph{Proceedings of the 54th Annual Meeting
  of the Association for Computational Linguistics (Volume 1: Long Papers)},
  Berlin, Germany, Aug. 2016, pp. 1715--1725. [Online]. Available:
  \url{https://www.aclweb.org/anthology/P16-1162}
\BIBentrySTDinterwordspacing

\bibitem{Kudo18subword}
T.~Kudo, ``Subword regularization: Improving neural network translation models
  with multiple subword candidates,'' in \emph{Proceedings of the 56th Annual
  Meeting of the Association for Computational Linguistics (Volume 1: Long
  Papers)}.\hskip 1em plus 0.5em minus 0.4em\relax Melbourne, Australia:
  Association for Computational Linguistics, 2018, pp. 66--75.

\bibitem{Kreutzer18learning}
\BIBentryALTinterwordspacing
J.~Kreutzer and A.~Sokolov, ``Learning to segment inputs for {NMT} favors
  character-level processing,'' in \emph{Proceedings of the international
  workshop on spoken language processing}, ser. IWSLT'18, Bruges, Belgium,
  2018. [Online]. Available: \url{http://arxiv.org/abs/1810.01480}
\BIBentrySTDinterwordspacing

\bibitem{Bansal18lowresource}
S.~Bansal, H.~Kamper, K.~Livescu, A.~Lopez, and S.~Goldwater,
  ``\BIBforeignlanguage{en}{Low-{{Resource Speech}}-to-{{Text Translation}}},''
  in \emph{\BIBforeignlanguage{en}{Interspeech 2018}}.\hskip 1em plus 0.5em
  minus 0.4em\relax {ISCA}, Sept. 2018, pp. 1298--1302.

\bibitem{Bansal18pretraining}
\BIBentryALTinterwordspacing
------, ``Pre-training on high-resource speech recognition improves
  low-resource speech-to-text translation,'' in \emph{Proceedings of the 2019
  Conference of the North {A}merican Chapter of the Association for
  Computational Linguistics: Human Language Technologies, Volume 1 (Long and
  Short Papers)}.\hskip 1em plus 0.5em minus 0.4em\relax Minneapolis,
  Minnesota: Association for Computational Linguistics, June 2019, pp. 58--68.
  [Online]. Available: \url{https://www.aclweb.org/anthology/N19-1006}
\BIBentrySTDinterwordspacing

\bibitem{Anastasopoulos18leveraging}
\BIBentryALTinterwordspacing
A.~Anastasopoulos and D.~Chiang, ``Leveraging translations for speech
  transcription in low-resource settings,'' in \emph{Proc. Interspeech 2018},
  2018, pp. 1279--1283. [Online]. Available:
  \url{http://dx.doi.org/10.21437/Interspeech.2018-2162}
\BIBentrySTDinterwordspacing

\bibitem{Anastasopoulos18tied}
------, ``Tied {{Multitask Learning}} for {{Neural Speech Translation}},'' in
  \emph{Proceedings of the 2018 {{Conference}} of the {{North American
  Chapter}} of the {{Association}} for {{Computational Linguistics}}: {{Human
  Language Technologies}}, {{Volume}} 1 ({{Long Papers}})}.\hskip 1em plus
  0.5em minus 0.4em\relax {New Orleans, Louisiana}: {Association for
  Computational Linguistics}, June 2018, pp. 82--91.

\bibitem{Adams18evaluating}
\BIBentryALTinterwordspacing
O.~Adams, T.~Cohn, G.~Neubig, H.~Cruz, S.~Bird, and A.~Michaud, ``Evaluation
  phonemic transcription of low-resource tonal languages for language
  documentation,'' in \emph{Proceedings of the Eleventh International
  Conference on Language Resources and Evaluation, {LREC} 2018, Miyazaki,
  Japan, May 7-12, 2018.}, 2018. [Online]. Available:
  \url{http://www.lrec-conf.org/proceedings/lrec2018/summaries/490.html}
\BIBentrySTDinterwordspacing

\bibitem{Godard18adaptor}
P.~Godard, L.~Besacier, F.~Yvon, M.~Adda-Decker, G.~Adda, H.~Maynard, and
  A.~Rialland, ``Adaptor grammars for the linguist: Word segmentation
  experiments for very low-resource languages,'' in \emph{Proceedings of the
  Fifteenth Workshop on Computational Research in Phonetics, Phonology, and
  Morphology}.\hskip 1em plus 0.5em minus 0.4em\relax Brussels, Belgium:
  Association for Computational Linguistics, October 2018, pp. 32--42.

\end{thebibliography}

\appendix
\section{Appendix}
\begin{figure*}[htpb]
  \centering
  \includegraphics[width=1.0\textwidth]{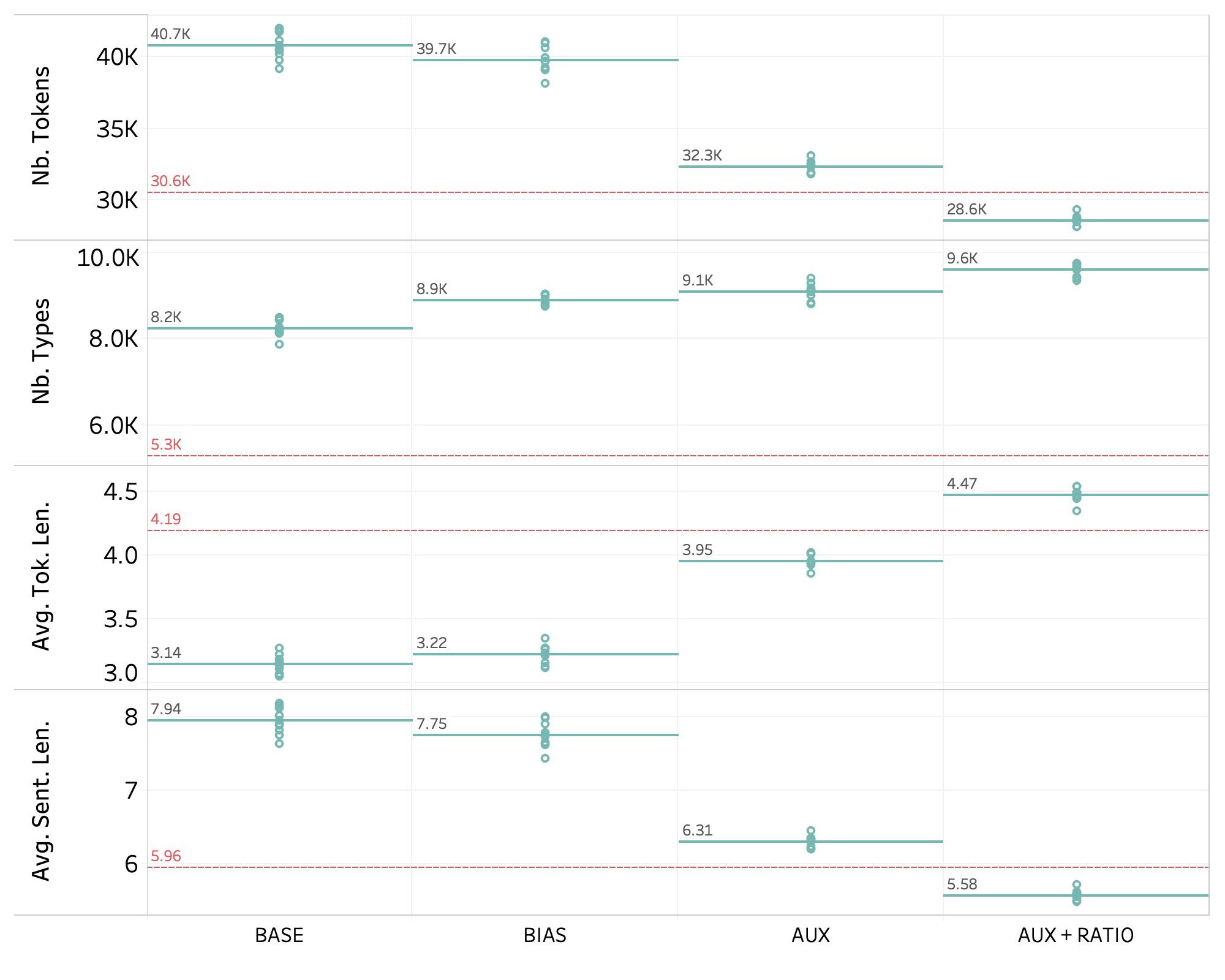}
  \caption[Statistics on segmentations produced by methods \base{}, \bias{}, \aux{}, and \auxratio{}, on the \mboshi{} 5K corpus for: number of tokens, types, average token length (in characters), average sentence lengths (in tokens).]{Statistics on segmentations produced by methods \base{}, \bias{}, \aux{}, and \auxratio{}, on the \mboshi{} 5K corpus: number of tokens, types, average token length (in characters), average sentence lengths (in tokens). Solid (teal-colored) lines correspond to average values (10 runs). Dashed (red) lines indicate the ground-truth values in the \mboshi{} 5K corpus.}
  \label{fig:att:base_ext_stats}
\end{figure*}



\end{document}